\documentclass[11pt,a4paper]{article}
\usepackage[hyperref]{acl2021}
\usepackage{times}
\usepackage{latexsym}

\usepackage{microtype}

\aclfinalcopy 

\usepackage{booktabs}
\usepackage{graphicx}
\usepackage{breakurl}
\usepackage{CJKutf8}
\newcommand{\ja}[1]{\begin{CJK}{UTF8}{min}#1\end{CJK}}
\newcommand{\minisection}[1]{\noindent{\bf {#1}}}

\title{Zero-pronoun Data Augmentation for Japanese-to-English Translation}

\author{Ryokan Ri, Toshiaki Nakazawa and Yoshimasa Tsuruoka\\
  The University of Tokyo \\
  7-3-1 Hongo, Bunkyo-ku, Tokyo, Japan \\
  \texttt{\{li0123, nakazawa, tsuruoka\}@logos.t.u-tokyo.ac.jp} \\}

\date{}

\begin{document}
\maketitle
\begin{abstract}
For Japanese-to-English translation, zero pronouns in Japanese pose a challenge, since the model needs to infer and produce the corresponding pronoun in the target side of the English sentence.
However, although fully resolving zero pronouns often needs discourse context, in some cases, the local context within a sentence gives clues to the inference of the zero pronoun.
In this study, we propose a data augmentation method that provides additional training signals for the translation model to learn correlations between local context and zero pronouns.
We show that the proposed method significantly improves the accuracy of zero pronoun translation with machine translation experiments in the conversational domain.

\end{abstract}

\section{Introduction}
While neural machine translation (NMT) has demonstrated high performance in single-sentence translation, it is still challenging to handle linguistic phenomena involving discourse contexts.
One such issue is the translation of {\it zero pronouns} (ZP) in Japanese-to-English translation.
In Japanese, subjects and objects are often omitted when the listener can infer them from the context. However, when translating them into English, the omitted words must be explicitly translated in most cases. For example, in the following sentence, the subject omitted in Japanese is the first person, and \textit{I} has to be output in English.

\begin{table}[h]
  \centering
  \begin{tabular}{ll}
  \ja{うなぎが}             & \ja{食べたいな}    \\
  unagi-ga         & tabe-tai-na              \\
  eel-\texttt{OBJ}          & eat-want-\texttt{PARTICLE}  \\
  \multicolumn{2}{l}{I feel like eating eel.}
\end{tabular}
\end{table}

The prediction of ZPs, essentially, requires understanding the topic and old information in the discourse, or referring to the world knowledge. On the other hand, linguistic information within the sentence may provide some clues \citep{kudo-etal-2015-anlp}.
For example, in the sentence above, the auxiliary verb \ja{たい} ({\it want}) suggests that the sentence expresses a subjective statement and thus the missing pronoun is the first person. Here we refer to such information as {\it local context}.

Correlations between local context and ZPs can be learned by the standard single-sentence neural machine translation, but it may not be possible under low-resource conditions. For example, the translation of conversations, which usually contain a large number of ZPs, is currently one of the under-resourced domains.

\begin{figure}[t]
\centering
\includegraphics[width=7.0cm]{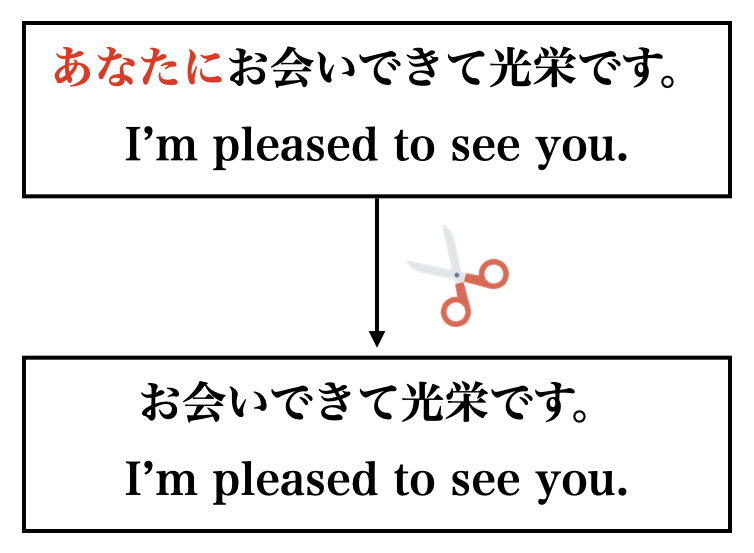}
\caption{The proposed method: ZP data augmentation}
\label{fig:zu}
\end{figure}

To address this problem, we propose {\bf zero pronoun data augmentation} to facilitate learning correlations between local context and ZPs (Figure \ref{fig:zu}).
We augment the training data by deleting personal pronouns in the source Japanese sentence.
This creates parallel data that include ZPs and provides additional training signals to learn to predict ZPs.
Our method is simple yet effective: it does not require any modification to the model architecture nor additional computation at inference time, but significantly improves the accuracy of the ZP translation.

\section{Related Work}
\subsection{Contextual Neural Machine Translation}
As the quality of single-sentence machine translation has improved dramatically with the advent of neural machine translation \citep{NIPS2014_a14ac55a,NIPS2017_3f5ee243}, translation models that take wider contexts into account have seen a surge of interest \citep{Jean2017DoesNM,bawden-etal-2018-evaluating,voita-etal-2019-good,voita-etal-2019-context,ma-etal-2020-simple,saunders-etal-2020-using}.
In contrast to the studies trying to incorporate information outside the sentence, in this work, we propose a method to improve zero-pronoun translation by only considering the information within the sentence, but we also explore the effect of combining our method with a contextual machine translation model.

\subsection{ZP Resolution in Japanese}
In some languages, pronouns are sometimes omitted when they are inferable from the context. Such languages are called pro-drop languages and the omitted pronouns are called ZPs.

The translation of ZPs poses a challenge when the corresponding pronoun is syntactically required on the target language side: the model has to infer the omitted pronoun.
The task of identifying the omitted pronouns is called ZP resolution and for Japanese, this has been a long-standing problem \citep{isozaki-hirao-2003-japanese,sasano-etal-2008-fully,imamura-etal-2009-discriminative,shibata-kurohashi-2018-entity}.
Japanese is one of the most difficult languages because Japanese words usually do not have any inflectional forms that depend on the omitted pronoun, unlike other pro-drop languages such as Portuguese and Spanish in which ZPs can be inferred from the grammatical case of other words.

Still, Japanese sentences sometimes contain expressions indicative of the missing pronoun. For example,  Japanese honorifics naturally indicate the subject is the second person.
In this work, we do not explicitly solve ZP resolution but let the translation model learn heuristic relations between ZPs and local context within the sentence \citep{hangyo-etal-2013-japanese,kudo-etal-2015-anlp} and produce appropriate English pronouns.

\subsection{ZPs in Translation}
In the context of statistical machine translation, Japanese ZPs are explicitly predicted by considering verbal semantic attributes \citep{nakaiwa-ikehara-1992-zero}, local context in the source and target sentence \citep{kudo-etal-2015-anlp}, and incorporated into the resulting translation.

On the other hand, in neural machine translation, the missing pronouns can be automatically inferred by the translation model because of the nature of end-to-end learning, although the correctness cannot be guaranteed. To improve the quality of ZP translation, previous studies have explored a multi-task approach with ZP prediction \citep{wang-etal-2016-novel,wang-etal-2019-one}.

In this study, we propose a ZP data augmentation method to provide additional training signals useful to correctly translate ZPs.

\section{Is Local Context Useful for Predicting Zero Pronouns?}
\label{section:analysis}

\begin{table*}[t]
\centering
\begin{tabular}{ccccccccccc}\toprule
                    & I    & you    & we    & they & he  & she & us  & them & him & her \\ \midrule
baseline            & 35.9 & 25.4   & 11.0  & 3.7  & 2.2 & 0.0 & 2.2 & 1.9  & 1.2 & 0.9 \\
logistic regression & 78.2 & 46.3   & 17.3  & 3.8  & 3.1 & 0.0 & 3.6 & 0.2  & 0.2 & 2.9 \\ \bottomrule
\end{tabular}
\caption{Recall scores of ZP predictions for each pronoun.}
\label{result:logistic}
\end{table*}

Our proposed method is based on the assumption that local context in Japanese sentences is useful for predicting ZPs.
We begin by analyzing to what extent ZPs can be inferred from local context, and what kind of local context is useful.

For the analysis, we use the Business Scene Dialogue Corpus \citep{rikters-etal-2019-designing}, which is a Japanese and English parallel corpus in the conversational domain.
Besides the published data, we also use the in-house version of the corpus, which amounts to a total of 104,961 sentence pairs.

\subsection{Identifying sentence pairs that contain ZPs.}
As the corpus does not contain annotations of ZPs, we first identify sentence pairs that contain zero pronouns.
We exploit the word alignment information from parallel sentences to detect ZPs. The specific procedure is as follows.

\begin{enumerate}
  \item We obtain the word alignments of the parallel data with \texttt{GIZA++}\footnote{\url{https://github.com/moses-smt/giza-pp}}. We use \texttt{Mecab}\footnote{\url{https://taku910.github.io/mecab/}} for Japanese word segmentation, \texttt{spaCy}\footnote{\url{https://spacy.io/}} for English.
  \item When a pronoun in an English sentence is associated with \texttt{NULL}, the pronoun in the English sentence is considered to correspond to a ZP in the Japanese sentence.
\end{enumerate}

\begin{figure}[t]
\centering
\includegraphics[width=7.0cm]{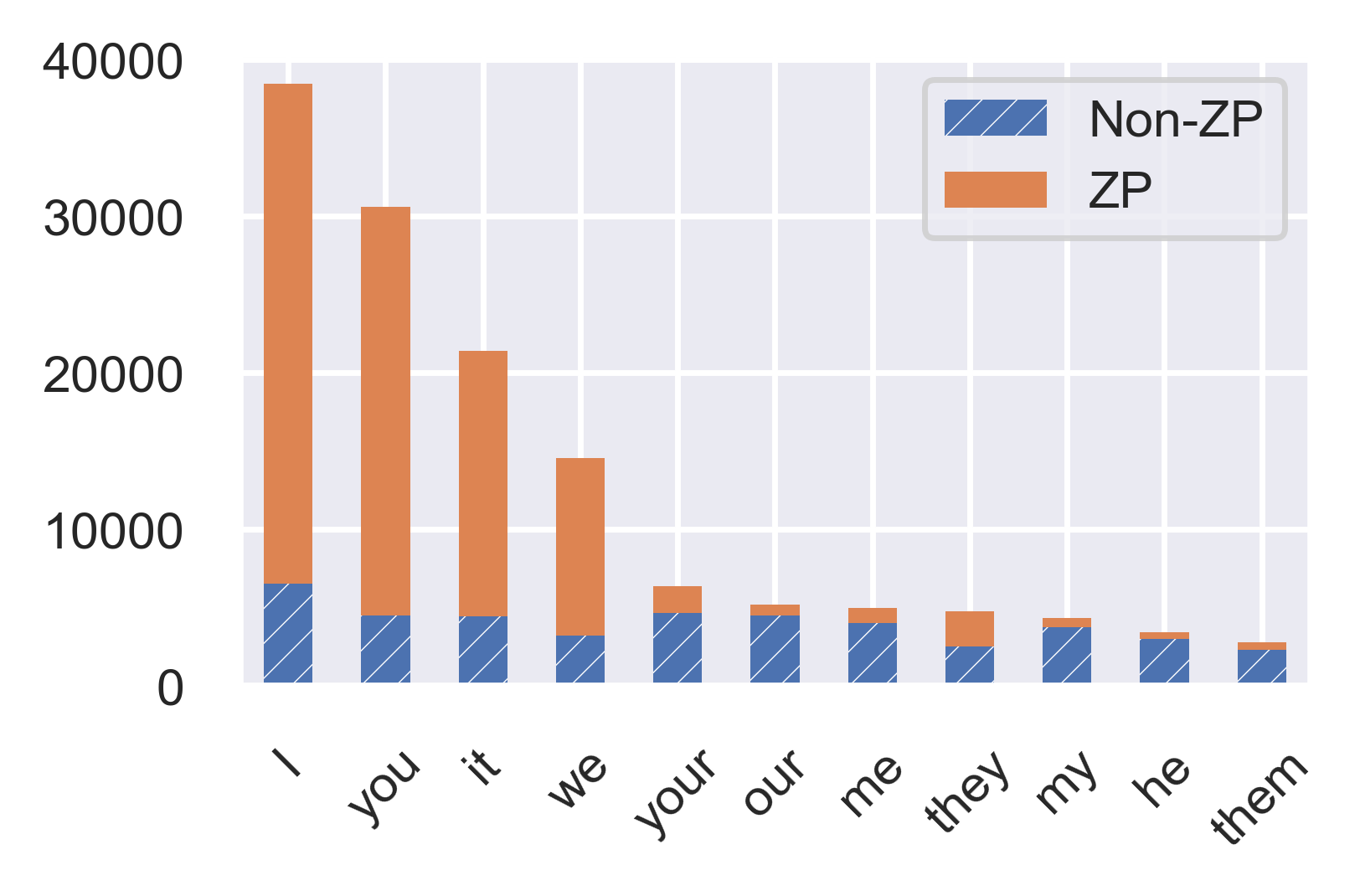}
\caption{The number of English pronouns in the analyzed data. ZP stands for those whose corresponding pronoun does not appear in the Japanese text.}
\label{fig:pronoun_count}
\end{figure}

The resulting number of pronouns is shown in Figure \ref{fig:pronoun_count}. It can be seen that in the conversational domain, the first person pronoun \textit{I} and the second person pronoun \textit{you} occur frequently and most of them ($80\%\sim$) are omitted in Japanese. More infrequent pronouns are less likely to be ZPs.

\subsection{Extracting local context that co-occurs with ZPs}
To associate the detected ZPs with local context in Japanese sentences, we extract the words that appear in their predicates.
We did not use a Japanese syntactic analyzer to detect ZPs but they are associated with the English pronouns by alignment. Therefore, we decided to exploit the alignment information to extract the predicates.
We extract the predicates of the English pronoun and the corresponding words in the Japanese sentence. Specifically, the following steps were taken.

\begin{enumerate}
  \setcounter{enumi}{2}
  \item We obtain the dependency tree of the English sentence with \texttt{spaCy} and extract the pronoun's head.
  \item The Japanese word aligned to the pronoun's head and its subsequent functional words \footnote{In this case, the function words are defined as words with one of the following part of speeches defined in \texttt{Mecab}: [``particle", ``auxiliary verb", ``symbol"].} are extracted as local context.
\end{enumerate}

\begin{table*}[t]
\centering
\begin{tabular}{cccc} \toprule
                  & 1to1        & 2to1        \\ \midrule
baseline          & 17.07$\pm$0.16 / 83.6$\pm$1.1 & 17.07$\pm$0.26 / 89.36$\pm$0.9 \\
baseline+pro\_aug & 17.07$\pm$0.19 / 92.32$\pm$1.8 & 17.11$\pm$0.23 / 92.17$\pm$1.1 \\ \bottomrule
\end{tabular}
\caption{Evaluation of the model with ZP data augmentation. The scores on the table are BLEU / ZP evaluation accuracy. The mean and standard deviation of five runs with different random seeds are reported.}
\label{result:mt}
\end{table*}

\subsection{Predicting ZPs from Local Context}
To investigate the extent to which ZPs can be predicted from local context, we conducted an analysis by training a logistic regression classifier \footnote{We use the implementation of the \texttt{scikit-learn} library with the default hyperparameters.}.
The classifier takes the unigrams, bi-grams, and trigrams extracted from local context in the Japanese sentence and predicts the associated pronoun in the English sentence.

The recall scores of each pronoun obtained with five-fold cross-validation are shown in Table \ref{result:logistic}. As a baseline, we adopt the score of random prediction according to the training distribution of pronouns.

One can see that the frequent pronouns such as \textit{I, you, we} can be predicted with significantly higher accuracy than the baseline when local context is used (around 6 to 43 points of improvement). In contrast, the other infrequent pronouns display similar or lower values compared to the baseline.
In summary, we can see that local context is predictive of the frequent pronouns but not for the infrequent ones.

To investigate what kind of local context is useful for prediction, for each output label ({\it i.e.}, pronoun) of the logistic regression classifier, we extracted the input features with higher values in the corresponding weights.
As a result, the following words are interpreted to be relevant.

\minisection{The first person singular \textit{I}}
verbs related to recognition (\ja{思う} (think), \ja{わかる} (understand), \ja{感じる} (feel)); humble words (\ja{申し上げる、存る}); and auxiliary verbs expressing desire (\ja{たい}).

\minisection{The second person singular \textit{you}}
suffixes expressing questions (\ja{かな？}, \ja{ました？}); speculations (\ja{でしょ}, \ja{だろ？}), honorifics (\ja{仰る}, \ja{いただける}).

\minisection{The first person plural \textit{we}}
obligations (\ja{なきゃ}, \ja{べき}), desire (\ja{たい}).

For the other pronouns, no local contexts were found to be interpretable as useful for prediction.

\section{ZP Data Augmentation}
\label{section:experiments}

In the previous section, we confirmed that local context is useful for predicting ZPs.
In this section, we examine the usefulness of ZP data augmentation for machine translation.

The method artificially creates training data containing ZPs by deleting pronouns in the source Japanese sentence along with the following particles.
The pronouns to be deleted are detected by string matching with manually created lists (Appendix \ref{appendix:list}).
The augmented data is supposed to provide useful training signals for learning correlations between ZPs and local context.

\subsection{Experimental Setups}
\minisection{Corpus}
We use the Document-aligned Japanese-English Conversation Parallel Corpus \citep{rikters-EtAl:2020:WMT}.
We also add an in-house conversational parallel corpus to the training data. The statistics of the corpus are shown in Table \ref{corpus:stat}.

\begin{table}[!h]
\centering
\begin{tabular}{cccc} \toprule
train    & train+pro\_aug     & dev       & test     \\ \midrule
\multicolumn{1}{c}{246,541} & \multicolumn{1}{c}{282,952} & \multicolumn{1}{c}{2,051} & \multicolumn{1}{c}{2,020} \\ \bottomrule
\end{tabular}
\caption{The number of sentences in the corpus.}
\label{corpus:stat}
\end{table}

\minisection{Model}
Transformer \citep{NIPS2017_3f5ee243} was used as the translation model. We adopt the hyperparameters recommended for the corpus of our size in \citet{araabi-monz-2020-optimizing} (Appendix \ref{appendix:hyperparameters}). In addition to the single-sentence translation, we also experimented with the 2to1 setting \citep{tiedemann-scherrer-2017-neural}, in which the previous sentence in the document is added to the input.

\minisection{Evaluation}
We evaluate the overall translation quality on the test set with BLEU \citep{papineni-etal-2002-bleu}.
We also conduct a targeted evaluation with the ZP evaluation dataset for Japanese-to-English translation \citep{shimazu-etal-2020-evaluation}. The ZP evaluation dataset contains 724 triples of a source sentence, a target sentence with a correct pronoun, and one with an incorrect pronoun.
To evaluate a translation model, we see if the model assigns a lower perplexity to the correct target sentence, and calculate the accuracy.

\subsection{Results}
The results of the experiment are shown in Table \ref{result:mt}.
We can observe that ZP data augmentation does not improve the BLEU score, but significantly improves the accuracy of ZP evaluation in both the 1to1 (83.6\% to 92.3\%) and 2to1 settings (89.3\% to 92.1\%). Our method yields a similar degree of improvement to the 2to1 setting in the ZP evaluation without any computational overhead at the inference time.

We also confirm that adding the previous context (2to1) does not improve BLEU but pronoun translation (83.6\% to 89.3\%), which conforms to observations in the previous study \citep{Jean2017DoesNM,shimazu-etal-2020-evaluation}.
However, this is not the case with the ZP data augmentation (92.3\% to 92.1\%).
We speculate that this is because longer inputs in the 2to1 setting make it more difficult for the model to find correlations between ZPs and local context.

\section{Conclusion}
To address the problem of zero pronoun translation, we proposed zero pronoun data augmentation.
Through the analysis with the Japanese-English conversational parallel corpus, we showed that zero pronouns in Japanese sentences can be predicted to some extent from local context within the sentence.
In the conversational translation experiment, we compared a translation model trained on the augmented data with the baseline and demonstrate that our method significantly improves the accuracy of zero pronoun translation.

Nevertheless, zero pronoun data augmentation does not solve the cases where the information necessary for zero pronoun translation exists outside the sentence.  Also, the analysis suggests that local context is useful for predicting frequent pronouns such as the first and second-person pronouns, but not for the third-person pronouns.
An interesting avenue for future work is to explicitly incorporate discourse-level contextual information such as topics or people involved in the conversation into the translation models.

\bibliographystyle{acl_natbib}
\bibliography{sections/reference}

\newpage
\onecolumn
\appendix
\section{The pronoun and particle list for pronoun data augmentation}
\label{appendix:list}
The deletion of pronouns was done by enumerating all combinations from the list of pronouns (Table \ref{pronoun_list}) and particles (Table \ref{aux_list}) and deleting strings that correspond to the pattern from the sentence.

\begin{table}[h]
\centering
\begin{tabular}{l|l} \toprule
 First person singular  & \ja{私, わたし, 僕, ぼく, 俺, おれ, わたくし, オレ, ウチ} \\
 First person plural   & \ja{我々, 僕ら, われわれ, 僕達, 僕たち, 私達} \\
 Second person singular & \ja{貴方, 貴女, あなた, お前, おまえ, 君, あんた} \\
 First person plural   & \ja{君たち, みなさま} \\
 Third person singular  & \ja{彼, 彼女, あいつ} \\
 Third person plural   & \ja{彼ら, 彼女ら, みんな, 皆, 皆んな, みなさん, 奴ら} \\
\bottomrule
\end{tabular}
\caption{The list of pronouns for pronoun deletion}
\label{pronoun_list}
\end{table}

\begin{table}[h]
\centering
\begin{tabular}{l|l} \toprule
  Nominative  &  \ja{は, が} \\
  Accusative  &  \ja{を} \\
  Dative      &  \ja{に} \\
  Possessive  &  \ja{の} \\
  Others      &  \ja{も, の方から, のほうから, の方に, のほうに, の方で}\\
              & \ja{のこと, の事, のほうで, から, 、} \\
\bottomrule
\end{tabular}
\caption{The list of particles for pronoun deletion}
\label{aux_list}
\end{table}

\section{Hyperparameters for the Machine Translation Experiment}
\label{appendix:hyperparameters}
We choose the hyperparameters of the Transformer model recommended in \cite{araabi-monz-2020-optimizing}.

\begin{table}[h]
\centering
\begin{tabular}{l|c} \toprule
 layers                        & 5 \\
 model size                    & 512 \\
 feed-forward dimension        & 2048 \\
 number of attention heads     & 4 \\
 encoder/decoder layer dropout & 0/0.1 \\
 src/tgt word dropout          & 0.2/0.2 \\
 label\_smoothing              & 0.3 \\
 optimizer                     & Adam with the Noam Learning rate schedule \\
 warmup steps                  & 8000 \\ \bottomrule
\end{tabular}
\caption{Hyperparameters for the Transformer model.}
\label{hyperparameters}
\end{table}

\end{document}